\documentclass{article}
\usepackage{microtype}
\usepackage{graphicx}
\usepackage{booktabs} 

\usepackage{listings}
\usepackage{color}
\definecolor{deepblue}{rgb}{0,0,0.5}
\definecolor{deepred}{rgb}{0.6,0,0}
\definecolor{deepgreen}{rgb}{0,0.5,0}

\definecolor{mygreen}{rgb}{0,0.6,0}
\definecolor{mygray}{rgb}{0.5,0.5,0.5}
\definecolor{mymauve}{rgb}{0.58,0,0.82}
\definecolor{myred}{rgb}{0.6,0.1,0.1}
\usepackage{hyperref}

\usepackage[accepted]{icml2019}

\newcommand{\toybox}{\textsc{ToyBox}}

\icmltitlerunning{\toybox{}}
\usepackage{enumitem}
\usepackage{multirow}
\usepackage{subcaption}
\usepackage{natbib}

\begin{document}

\twocolumn[
\icmltitle{\toybox{}: A Suite of Environments for \\Experimental Evaluation of Deep Reinforcement Learning}




\begin{icmlauthorlist}
\icmlauthor{Emma Tosch}{umass}
\icmlauthor{Kaleigh Clary}{umass}
\icmlauthor{John Foley}{smith}
\icmlauthor{David Jensen}{umass}
\end{icmlauthorlist}

\icmlaffiliation{umass}{University of Massachusetts Amherst, Amherst, USA}
\icmlaffiliation{smith}{Smith College, Northampton, USA}

\icmlcorrespondingauthor{Emma Tosch}{etosch@cs.umass.edu}

\icmlkeywords{Deep Reinforcement Learning, Benchmarks, Systems, Atari, Explainability, Evaluation}

\vskip 0.3in
]



\printAffiliationsAndNotice{}  


\begin{abstract}
Evaluation of deep reinforcement learning (RL) is inherently challenging.
In particular, learned policies are largely opaque, and hypotheses about the behavior of deep RL agents are difficult to test in black-box environments.
Considerable effort has gone into addressing opacity, but almost no effort has been devoted to producing high-quality environments for experimental evaluation of agent behavior.
We present \toybox{}, a new high-performance, open-source* subset of Atari environments re-designed for the experimental evaluation of deep RL.
We show that \toybox{} enables a wide range of experiments and analyses that are impossible in other environments.
\\
{\small*\url{https://kdl-umass.github.io/Toybox/}}
\end{abstract}

\section{Introduction}
\label{sec:introduction}
Since DeepMind's 2015 Nature paper, the Arcade Learning Environment (ALE) has become the \emph{de facto} deep RL benchmark for new training algorithms~\cite{bellemare13arcade, mnih2015nature, machado17arcade}. ALE has several appealing qualities: humans learn to play Atari and become more skilled with experience, it is a ``real-world'' environment that was not originally constructed to evaluate RL methods, and it has greater complexity than prior environments (e.g., GridWorld, mountain car).


ALE has been used in several ways to evaluate the performance of deep RL agents. The vast majority of evaluations follow a version of the approach described by \citet{bellemare2012investigating}: first researchers choose network architectures and tune hyperparameters on small set of Atari games; then they train agents using those hyperparameters on new games, reporting the learning curves (or a statistic of a collection of those curves)~\cite{mnih2015nature, van2016deep, mnih2016asynchronous, hessel2018rainbow}. 

While ALE has enabled demonstration and evaluation of much more complex behaviors of deep RL agents, it presents challenges as a suite of evaluation environments for topics on the frontier of deep RL. 

\textit{Challenge: Limited variation within games.} Very little about individual games can be systematically altered, so ALE is poorly suited to testing how changes in the environment affect training and performance. New benchmarks such as OpenAI's \textit{Sonic the Hedgehog} emulator and \textit{CoinRun} inject environmental variation into the training schedule, while introducing train/test splits~\cite{nichol2018gotta, cobbe2018quantifying}. Similarly, \citet{zhang2018natural} suggest benchmarks that incorporate the kind of non-random noise found in nature. \citet{kansky2017schema} implemented Breakout variants in order to achieve variation for generalization. 

\textit{Challenge: No counterfactual evaluation.} Meanwhile, assertions about intelligent agent behavior remain untestable in the face of black-box evaluation environments. For example, ALE does not enable testing the conjecture that agents trained on Breakout learn to build tunnels \cite{mnih2015nature} or that they enter a tunneling mode \cite{greydanus2017visualizing}. No system currently permits experiments to answer counterfactual questions about agent behavior. 

\textbf{Contribution.} We propose \toybox{}, a suite of high-performance and highly parameterized 
Atari-like environments designed for the purpose of experimentation. We demonstrate that the \toybox{} implementations of three Atari 2600 games achieve similar performance to their ALE counterparts across three deep RL algorithms.  We demonstrate that \toybox{} enables a range of post-training analyses not previously possible, and we show that \toybox{} is orthogonal to concurrent efforts in deep RL to address issues of robustness, generalization, and reproducible evaluation. 

\textbf{Organization.} The rest of the paper is organized as follows: Section~\ref{sec:toybox} introduces the \toybox{} design and its functional capabilities. Section~\ref{sec:evaluation} describes our evaluation, including performance and fidelity testing against the ALE. Section~\ref{sec:case-studies} describes four behavioral tests we present as case studies. Related work not otherwise addressed can be found in Section~\ref{sec:related}.  Section~\ref{sec:discussion} discusses \toybox{} applications beyond the scope of this paper. We conclude in Section~\ref{sec:conclusion}.

\section{\toybox{}: System Design}
\label{sec:toybox}

\toybox{} is a high-performance, highly parameterized suite of Atari-like games implemented in Rust, with bindings to Python. The suite currently contains three games:  Breakout, Amidar, and Space Invaders.  We chose these games for diversity of genre (paddle-based, maze, and shooter, respectively) and likely familiarity to readers.\footnote{Although Amidar may not be familiar, it is very similar to PacMan, but with simpler rules.} 

\textbf{Software requirements.} Atari 2600 games were designed for human players. For \toybox{}, the primary user is a reinforcement learning algorithm, and we expect machine learning researchers to be able to customize gameplay. To that end, we developed \toybox{} to meet the following set of software requirements
\begin{enumerate}[label=\textbf{R\arabic*}, leftmargin=*]
    \item\label{req:cpu}\toybox{} should only leverage the CPU, even for graphical tasks. Although modern games leverage the GPU for faster rendering, we expect the machine learning libraries to be using the GPU, and so we wish to create our screen images using CPU-only. 
    
    \item\label{req:efficiency}\toybox{} should be at least as efficient as the Stella-emulated version of the game. Since reinforcement learning algorithms require millions of frames of training data, we must be able to simulate and render millions of frames in order to enable efficient use of computation resources for learning.
    
    \item\label{req:data-driven}\toybox{} should provide for data-driven user customization. Changing the bricks in Breakout, the board in Amidar, or the alien configuration in Space Invaders should not require re-compilation of the core game code, nor should it require the ability to write Rust.
    
    \item\label{req:gym}\toybox{} should be accessible through OpenAI Gym, which is a Python API. Furthermore, \toybox{} should be usable as a drop-in replacement for the analogous ALE environment. 
    
\end{enumerate}
\begin{figure}[!h]
    \centering
    \includegraphics[width=\linewidth]{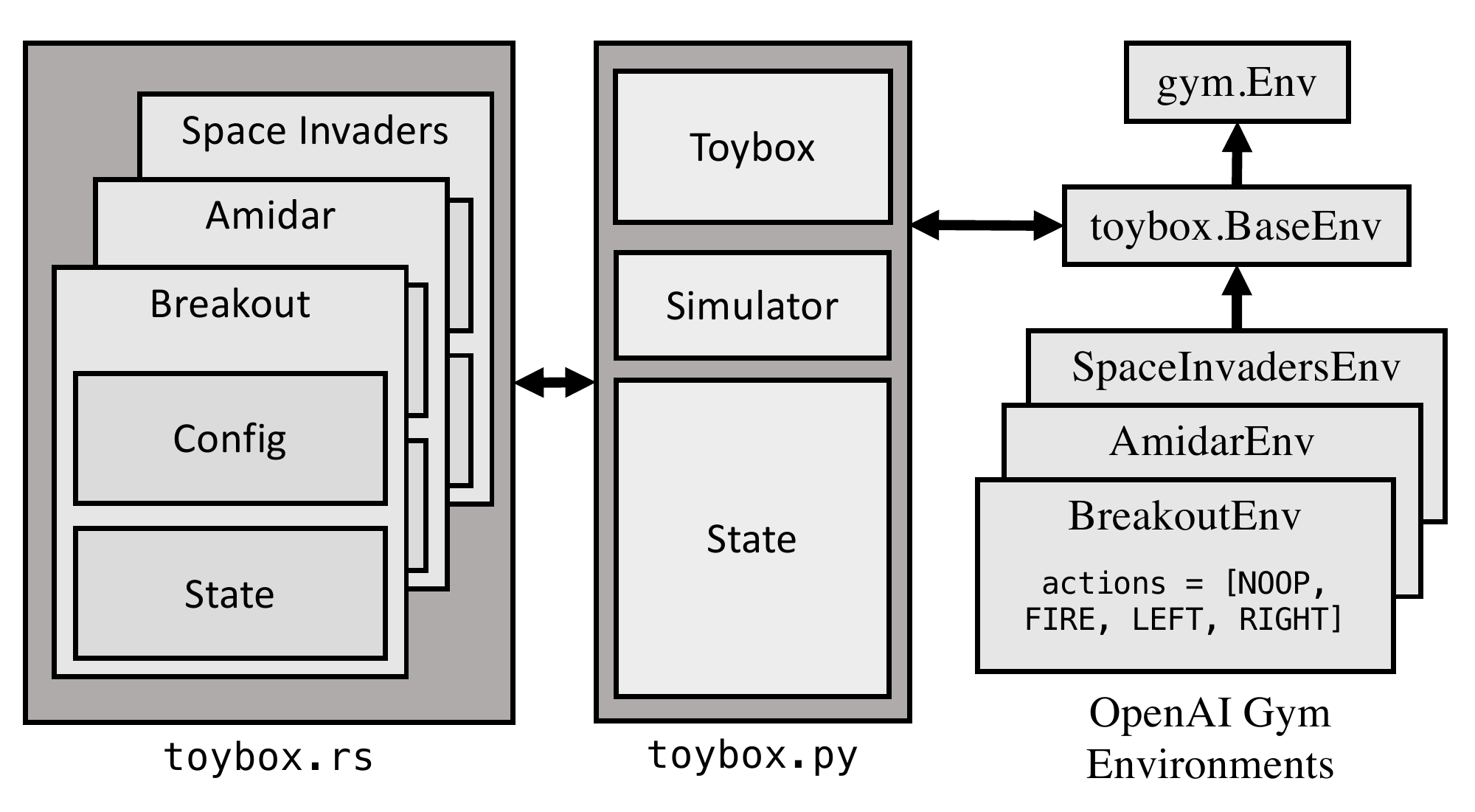}
    \caption{The \toybox{} architecture.}
    \label{fig:toybox-system}
\end{figure}

\begin{figure}[!t]
    \centering
    \includegraphics[width=0.8\columnwidth]{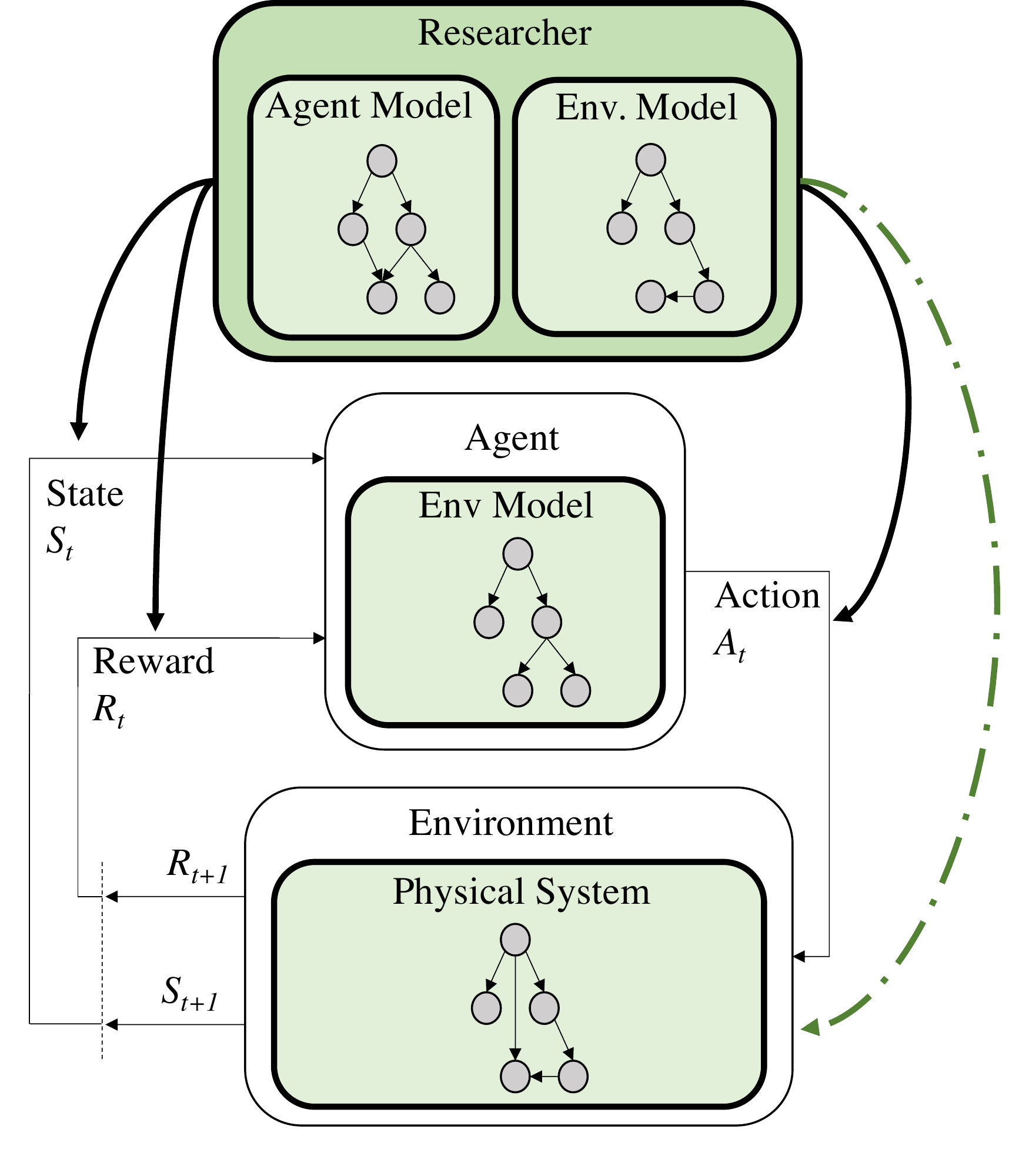}
    \caption{A traditional RL diagram, augmented with features required for counterfactual reasoning (shaded above). There is a physical system that governs behavior within the environment that may only be partially known. The researcher has a model of the environment's physics and, when attempting to explain agent behavior, must implicitly form a model of the agent's model. Arrows emanating from the researcher point to elements of the system where the researcher may intervene. The dotted line from the researcher to the environment is possible in \toybox{}, but is not typically considered.}
    \label{fig:rl-system}
\end{figure}

\textbf{Architecture.} Figure~\ref{fig:toybox-system} depicts the \toybox{} architecture. The game logic is written in Rust. Every game implements two core structs: \texttt{Config} and \texttt{State}. The \texttt{Config} struct contains data that we would generally only expect to be initialized at the start of an episode (i.e., a game, which may include multiple lives). The \texttt{State} struct contains data that may change between frames. At any point during execution, a \toybox{} game can be paused, state exported and modified, and resumed with the new state. 


\textbf{Aside: Why Atari?} Given the range of open problems in deep RL, creating a new ALE-like system may not seem like an effective way to facilitate cutting-edge deep RL research. However, there are still many poorly understood properties of Atari games and agent behavior on them.

There are many axes of complexity in the reinforcement learning environment: required planning horizon, reward function assignment, number of actions available, environment stochasticity, and  state representation are all different components affecting the total complexity of the environment, each presenting unique challenges for policy learning:

\pagebreak
\begin{itemize}[leftmargin=*]
  \item \textbf{Planning Horizon} Atari environments requiring temporally extended planning strategies for success, as in Frostbite, still present challenges to deep reinforcement learning algorithms ~\cite{jiang2015dependence, lake2017building, farquhariclr218}. 
    \item \textbf{Reward Function} Environments with sparse reward functions are another sticking point for RL, having only recently seen significant progress on games in this category~\cite{goexplore, salimans2018,burda2018exploration}. Atari examples include Pitfall and Montezuma's Revenge. 
    \item \textbf{State Representation} Operating over pixel representations like Atari games means the agent experiences a huge state space, but the underlying objects defining those sensory representations can often be represented more compactly~\cite{guestrin2003generalizing, littman2008oomdp, kansky2017schema}. Models trained on a compact representation of state can train more quickly~\cite{brunskill2018strategic, melnik2018modularization}, but the underlying environment dynamics have not changed.
    \item \textbf{Action Space Complexity}  Increasing the size of the action space can quickly lead to intractable computation for Q-value or policy function approximation, especially when that approximation is computationally expensive as in deep RL ~\cite{dulac2015deep}. Atari games available in ALE allow 4-18 actions ~\cite{mnih2013playing}. While this may seem small, even environments with as few as 10 actions can present challenges to efficient learning~\cite{dulac2012fast}. 
    \item \textbf{Environment Stochasticity} Previous work has shown that environments with higher stochasticity can be more difficult for some types of RL algorithms to learn~\cite{henderson2017deep}. Atari games are deterministic, but \toybox{} enables Atari-like environments with parameterized stochasticity.
    
\end{itemize}



Despite these open challenges, it has been argued that Atari's environments are not sufficiently complex to evaluate reinforcement learning agents because the source code is small~\cite{zhang2018natural}. However, source code size as minimum description length is a poor proxy for environment complexity. As ~\citet{raghu2017can} have shown, Erd\H{o}s-Selfridge-Spencer games can be represented quite compactly, simply requiring the assignment of two parameters, but represent a large combinatorial space of potential games for evaluating RL algorithms. 

Ultimately, each of these axes of complexity is observed through the agent's interaction with the environment. Figure~\ref{fig:rl-system} depicts the traditional RL diagram, overlaid with the interventions a researcher may make, as well as the models that inform both the researcher's and the agent's decision-making. The solid arrows represent current avenues for experimentation used by deep RL researchers to evaluate models: sensory perception of state, action selection manipulations, and reward function definition. 

Without intervention or introspection of the environment, researchers must use observational data of agent behavior to reason about experimental results. \toybox{} enables new methodology for experimentation in deep RL. Ultimately, we mimic Atari games because there is a font of untapped research questions related to testing and explaining the behavior of deep RL agents. We chose our initial set of games to establish that there are surprising results on even seemingly ``solved'' games such as Breakout and Space Invaders.  



 %


\section{Evaluation}
\label{sec:evaluation}
\textbf{Performance.} 
We achieved \ref{req:cpu} by designing a simple CPU graphics library; we demonstrate \toybox{} efficiency (\ref{req:efficiency}) in Table~\ref{tab:performance}. Note that \toybox{} permits researchers to process games entirely in grayscale and achieve substantial additional performance gains.  However, since this is not a feature offered in ALE, we only compared against the \toybox{} RGB(A) rendering.

\begin{table}[t!]
    \caption{\textbf{\toybox{} vs ALE performance}. Thousand frames per second (kFPS) on a MacBook Air (OSX Version 10.13.6) with a 1.6 GHz Intel Core i5 processor having 4 logical cores. Rates are averaged over 30 trials of 10~000 steps and reported to two significant digits, with standard error. We consistently observed an approximately 95\% slowdown when interacting with both ALE (C++) and \toybox{} (Rust) via OpenAI Gym. All benchmarks are run from CPython 3.5 and include FFI overhead (via \texttt{atari-py} for ALE).}
    \centering
    \begin{tabular}{p{0.18\linewidth}p{0.15\linewidth}*{2}{p{0.22\linewidth}}}
&& Raw kFPS & Gym kFPS\\
        \cline{3-4}
        \multirow{2}{*}{{Breakout}} & {ALE} & 52 (1.3) & 3.4 (0.065) \\
        &{\toybox{}} & 230 (5.4) & 7.2 (0.23)\\
        \cline{2-4}
        \multirow{2}{*}{{Amidar}} & ALE & 61 (2.9) & 3.0 (0.083)\\
        & \toybox{} & 250 (2.3) & 6.0 (0.112)\\
        \cline{2-4}
        \multirow{2}{*}{\parbox[t]{4cm}{Space\\Invaders}} & ALE & 55 (1.3) & 3.9 (0.072)\\
        & \toybox{} & 120 (3.4) & 5.2 (0.082)
     \end{tabular}
    \label{tab:performance}
\end{table}

\begin{figure*}[t!]
  \begin{minipage}[b]{.27\linewidth}
    \centering
    \includegraphics[width=0.9\linewidth]{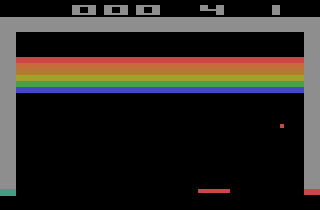}
    \includegraphics[width=0.9\linewidth]{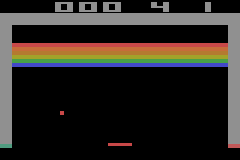}
    \subcaption{\small Upper: ALE; lower: \toybox{}.}
  \end{minipage}\quad%
  \begin{minipage}[b]{0.4\linewidth}
    \centering
      \begin{minipage}[b]{0.49\linewidth}
        \includegraphics[width=\linewidth]{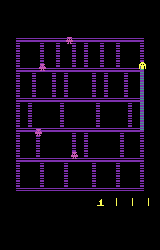}
      \end{minipage}~%
      \begin{minipage}[b]{0.49\linewidth}
        \includegraphics[width=\linewidth]{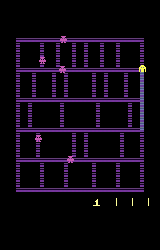}
      \end{minipage}
      \subcaption{\small  Left: ALE; right: \toybox{}.}
  \end{minipage}\quad%
  \begin{minipage}[b]{.27\linewidth}
    \centering
    \includegraphics[width=0.9\linewidth]{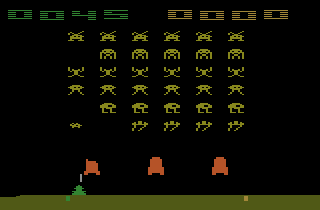}
    \includegraphics[width=0.9\linewidth]{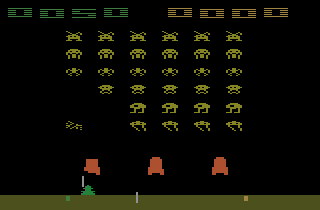}
    \subcaption{\small Upper: ALE; lower: \toybox{}.}
    \label{fig:space-invaders}
\end{minipage}
  \caption{Side-by-side comparisons of screen shots from ALE and \toybox{} Atari games. Each game represents a different deterministic action trace, but traces are the same between ALE and \toybox{}. \toybox{} implementations of Breakout and Space Invaders have nondeterministic elements. Amidar is deterministic in both ALE and \toybox{}. Due to idiosyncrasies at the start of ALE Amidar gameplay, the frames are not from identical points in the action trace; frames for Breakout and Space Invaders are. Note that for Amidar ALE appears to be missing an enemy, which is due to Atari only rendering a subset of sprites each frame due to computational constraints.}
  \label{fig:screenshots}
\end{figure*}

\textbf{Fidelity.} Figure~\ref{fig:screenshots} depicts frames at roughly equivalent points in the execution of a fixed action trace. If \toybox{} perfectly reproduced the games in ALE, the frames would be exactly the same. Three factors prevent exact replication of games: (1) Atari 2600 game source code is not available, (2) there are no formal specifications and few informal specifications of games,\footnote{Some informal specifications contain errors (e.g., the Atari manual for Breakout refers to a row of bricks that does not exist).} and (3) inferring arbitrarily complex programs from data is extremely challenging~\cite{raychev2016learning}. 
Therefore, comparing frames of program traces is not a sufficient measure of how closely we have approximated ALE games.

A human study could help assess whether the two implementations are are sufficiently alike. While we did solicit feedback from Atari aficionados during development (via a playable interface),
we did not view human perception of equivalence as a sufficient measure of fidelity. Human players rely on unique problem-solving capabilities that deep RL agents have not yet achieved, while deep networks are undeterred by the kind of noise that can confuse humans  ~\cite{adversarial,dubey2018investigating}. Instead, we focused on tuning our environments to produce comparable post-training agent performance. 

\begin{figure}
    \centering
    \includegraphics[width=\columnwidth]{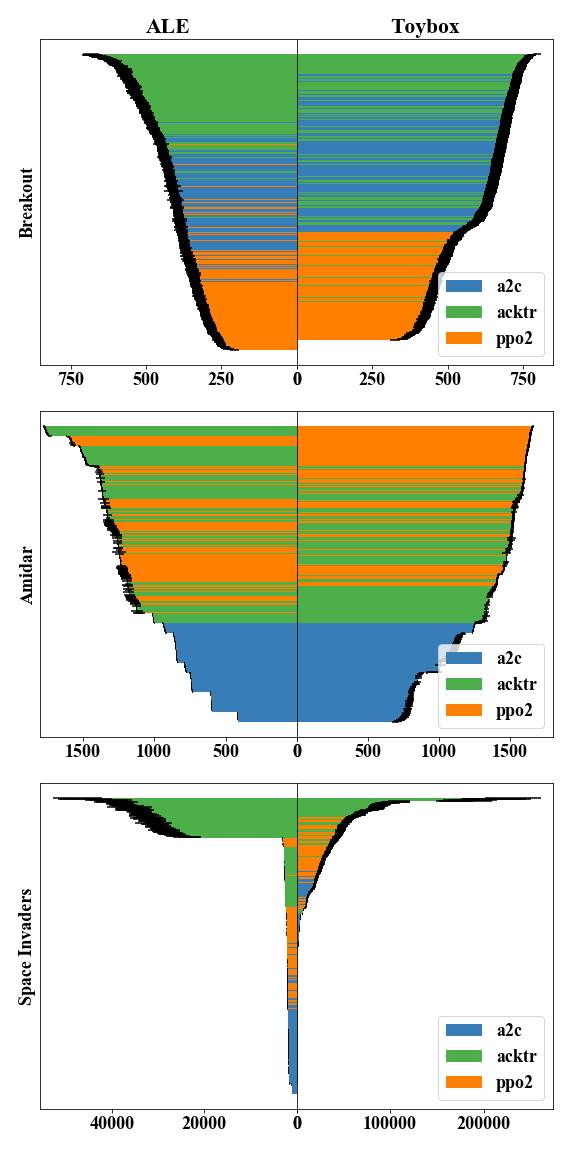}
    \caption{Rankings across 300 model replicates per model, per game, with standard error. Each horizontal bar is the average performance for a trained model, evaluated over 30 games. Stridations should be similar in both environments. \textbf{Breakout}: One training seed led an agent trained using \texttt{acktr} to have abysmally poor performance in every trial (i.e., all average scores below 2). \textbf{Amidar}: The lack of within-game variation is reflected in the short error bars and similar rankings between backends. \textbf{Space Invaders}: \toybox{} only implements the first level of Space Invaders. Since there is currently no way to detect levels within ALE, we let both agents play indefinitely. However, because Space Invaders levels vary considerably, performance is not comparable. For this reason we use different scales for the score (x-axis).}
    \label{fig:ranking}
\end{figure}

\textbf{Methodology.} We used three off-the-shelf implementations of training algorithms with default parameter settings for 5e7 steps from OpenAI Baselines~\cite{baselines}: \texttt{a2c}~\cite{mnih2016asynchronous},
\texttt{acktr}~\cite{wu2017scalable}, and
\texttt{ppo2}~\cite{schulman2017proximal}. Due to issues with variability across agents and environments~\cite{henderson2017deep, clary2018variability, jordan2018benchmarks}, we trained ten replicates for each of these training algorithms, differentiated by their random seed. Since there are various other uncontrolled sources of randomness, we evaluated each of these thirty agents per game using thirty unique random seeds over thirty games. Figure~\ref{fig:ranking} depicts our results: we find that agents achieve sufficiently similar performance in each analogous environment, and have roughly equivalent rankings (idiosyncrasies discussed in the in Fig.~\ref{fig:ranking} caption).



\textbf{Findings.} \toybox{} is much faster than ALE and our ranking results in Fig.~\ref{fig:ranking} show that our current implementations are comparable to their ALE counterparts. We will continue to strive for fidelity over the course of game development. 

\section{Case Studies}
\label{sec:case-studies}
We demonstrate~(\ref{req:data-driven}: Customization) with four case studies, in which we test the post-training performance\footnote{The analyses in this paper focus on evaluations of post-training performance, but \toybox{} interventions can be applied at any time--including during training.} of agents according to some hypothesis about behavior. This sort of testing is useful for evaluating a single agent prior to deployment (i.e., acceptance testing) or as existential proof for behavior under counterfactual conditions. None of these tests are currently possible in ALE because they all rely on resuming gameplay from an arbitrary modified state. 
Furthermore, all experiments in this section can be expressed in fewer than 200 lines of code, in addition to the code required for loading up the OpenAI baselines models. We include a code snippet from one of our experiments in Figure~\ref{fig:snippet}.


\begin{figure}
    \centering
    \begin{minipage}[c]{.33\linewidth}
        \centering
        \includegraphics[trim={50 50 50 50}, clip, height=1.25in]{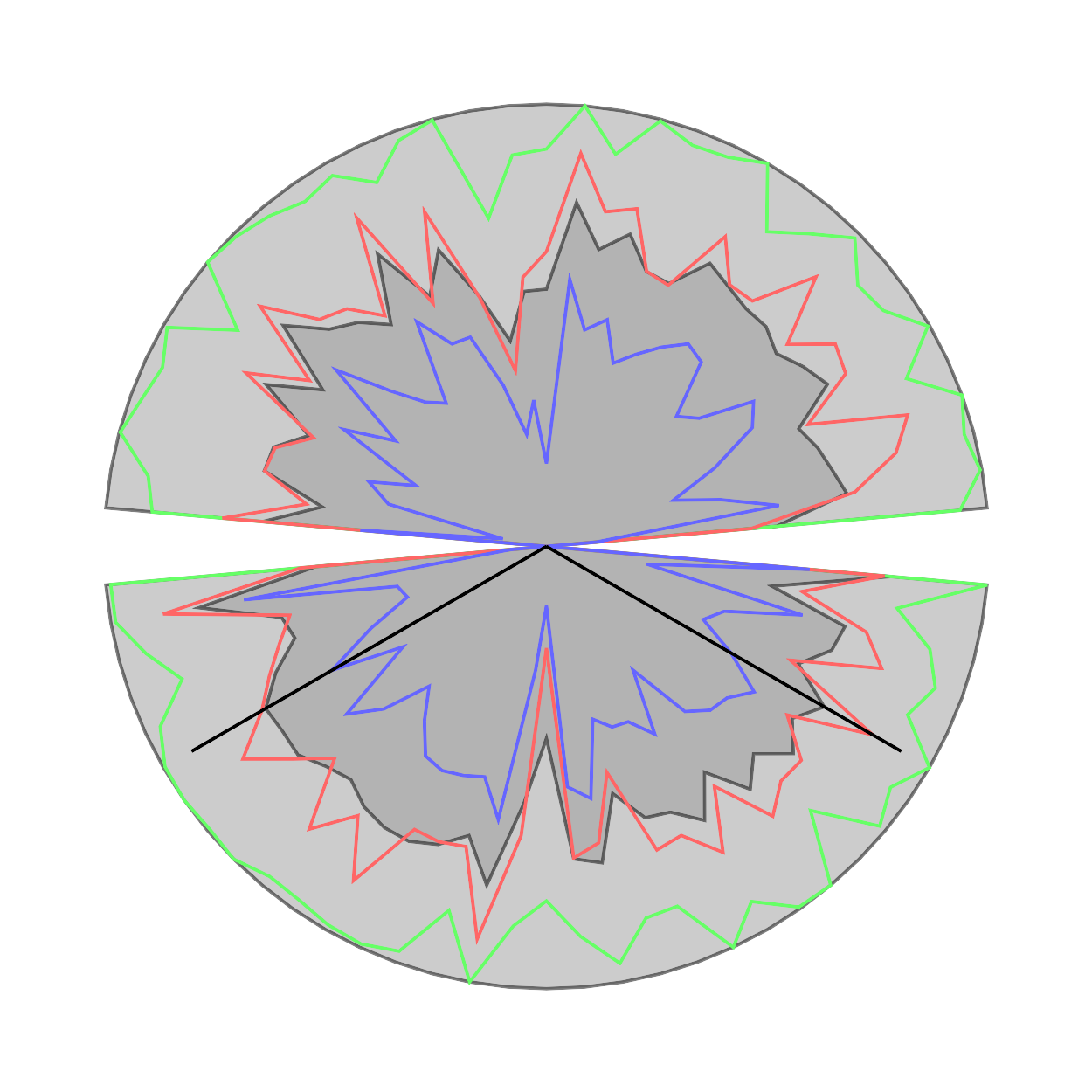}
        \subcaption{Polar starts}
        \label{fig:polar-starts}
    \end{minipage}~%
    \begin{minipage}[c]{0.66\linewidth}
        \centering
        \includegraphics[height=1.25in]{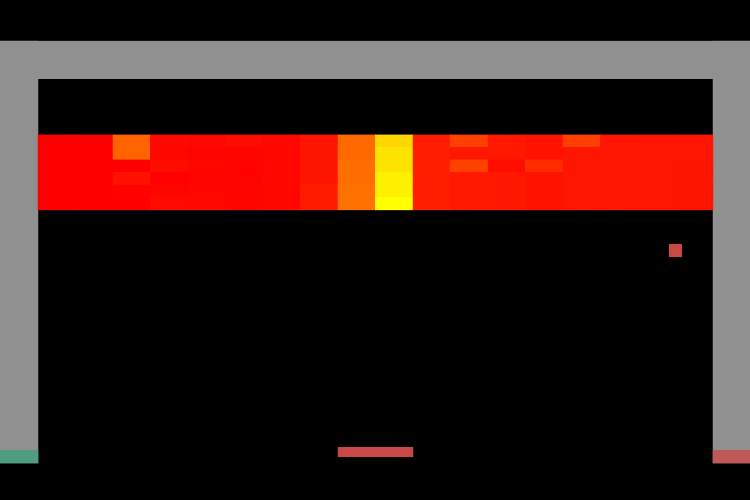}
        \subcaption{Tunneling}
        \label{fig:tunneling}
    \end{minipage}
    \label{fig:breakout}
    \caption{\textbf{Breakout case study.} Results from counterfactual queries for starting angles and tunnels. Both tests were run on an agent trained with the default OpenAI Baselines parameters for \texttt{ppo} for 5e7 steps, one life, and a 4 min. timeout. \textbf{(a)}:~The black lines indicate the starting angles seen during training, the light gray area the maximum score achieved from this starting angle, and the dark gray area represents the mean score achieved across trials. \textbf{(b)}:~Brick hue scaled according to the inverse of the median number of steps required to clear the particular brick in that test; bright-yellow represents fewer steps and red represents more steps (17-400).}
\end{figure}
\begin{figure*}[!h]
    \centering
    \begin{minipage}[c]{0.55\linewidth}
        \centering
    \includegraphics[width=\columnwidth]{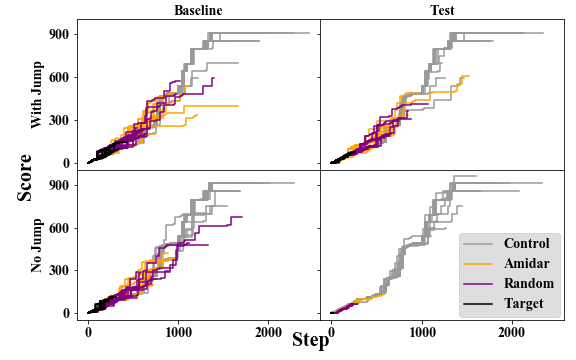}
        \subcaption{Accumulated game score of a single \texttt{a2c}-trained agent.}
        \label{fig:amidar-test-single}
        \end{minipage}\quad%
    \begin{minipage}[c]{0.4\linewidth}
        \centering
        \includegraphics[width=\linewidth, height=2.5in]{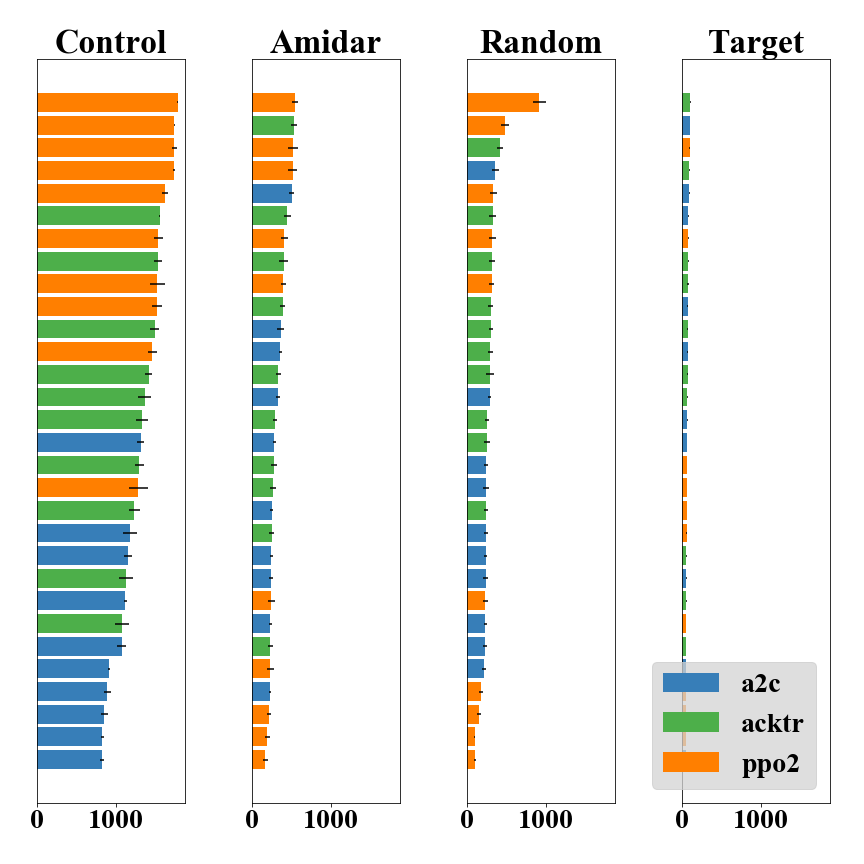}
        \subcaption{Ranking of 30 model replicates.}
    \label{fig:amidar-test-rank}
    \end{minipage}
\caption{\textbf{Amidar case study.} for four enemy movement protocols (``Control'' is a lookup table, ``Amidar'' is the ``Amidar movement,'' ''Random'' enemies move in a random direction at every junction, and ``Target'' causes enemies to pursue the player when it is in line of sight). \textbf{(a)}: \textit{Upper left}: baseline performance of the agent on each of the four protocols. \textit{Lower left}: baseline performance of the agent on each protocol without the ability to jump over enemies. \textit{Upper right}: ``Ganging up'' test, where all agents start close to the player. \textit{Lower right}: ``Ganging up'' test with no jump. \textbf{(b)}: Score ranking of 30 model replicates for the baseline condition with jumps (i.e., the upper left corner of the left graph) for each movement protocol.}
\end{figure*}

\paragraph{Breakout: Polar Angles.} An agent that has learned to play Breakout must have at least learned to hit the ball. Our first test manipulates the starting angle of the ball. 

We modified the start state to change the initial launch angle of the ball in 5$^\circ$ increments (72 configurations). Figure~\ref{fig:polar-starts} depicts the results. Note that the agent fails to achieve any score with horizontal ball angles: since Breakout has no gravity, balls simply bounce horizontally forever, never hitting any bricks or threatening the paddle. The agent also sometimes struggled with vertical angles. When we observed this behavior, the agent would keep the ball aligned perfectly in the center of the board, hitting it precisely in the center of the paddle, failing to make progress. This is an unexpected behavior that is entirely unlike human gameplay. In all, we found the agent to be resilient to starting angles, albeit with high variance. This suggests that an agent can be successful even with balls traveling at angles it may never have observed in training, a powerful recommendation for the training algorithms that produced such robust RL agents. 

\paragraph{Breakout: Tunneling.} One of the most promising behaviors observed in deep RL has been the apparent ability to learn higher-level strategies. Perhaps no high-level strategy has been written about more than ``tunneling'' in the game of Breakout, which happens when the player clears a column of bricks, causing the ball to bounce through the hole and onto the ceiling, clearing many bricks rapidly~\cite{mnih2015nature, greydanus2017visualizing}. One way to test whether an agent intentionally exploits tunneling is to give it a board with a nearly complete tunnel, save for a single brick, and test whether the agent can prioritize aiming at that single brick. 

For every brick, we removed all other bricks in the column, creating a nearly-completed tunnel. Figure~\ref{fig:tunneling} depicts the results: the value for each brick is the reciprocal of the median number of time steps before that brick was removed. 

If an agent were able to build tunnels, we would expect, for example, symmetry along one or both axes. Instead, we see that the agent clears one column in the center very quickly, the left adjacent column and some bricks in the upper left region a bit more slowly, and the remaining bricks take all about the same time. Observing agent gameplay, we saw the agents hit the ball to predictable locations, regardless of the board configuration. 

\begin{figure}
    \centering
    \footnotesize
    \begin{lstlisting}
# Fourth test: Target Player
# Delete Existing Enemies
config = toybox.get_config_json()
config['enemies'] = []
# Add Enemies that chase the player:
for i in range(5):
    config['enemies'].append({
        'EnemyTargetPlayer' : {
            'start' : starts[i],
            'start_dir': 'Right',
            'dir': 'Right',
            'player_seen': None
    }})
# Update game configuration
toybox.write_config_json(config)
# Run 30 trials with OpenAI Gym API
obs = env.reset()
for trial in range(n_trials):
    n_steps = 0
    done = False
    # until death or max_steps
    while n_steps < max_steps and not done:
        action = agent.step(obs)
        obs, _, death, info = env.step(action)
        done = death and not toybox.game_over()
        log_step(toybox)
        n_steps += 1
    log_after_episode(toybox)
    \end{lstlisting}
    \caption{Code snippet from our Amidar Protocol test. We build JSON config in Python and run it in \toybox{}.}
    \label{fig:snippet}
\end{figure}
\paragraph{Amidar: Enemy Protocol.} Suppose we would like to test whether an agent has learned to avoid adversarial elements of a game: e.g., the enemies in Amidar. To test this, we might drop the agent around the corner from an enemy, or position the enemies to ``gang up'' on the player, forcing the agent to move in a particular direction. 

This kind of intervention is only meaningful if enemy position is a function of current location. Observation led us to conclude that enemies move in fixed loops, likely implemented as lookup tables. This contrasts with ``Amidar movement'' is believed to dictate enemy behavior.\footnote{An enemy moves with a diagonal velocity, flipping the vertical direction when encountering the top or bottom of the board and horizontal direction when encountering the left or right edge (\url{https://en.wikipedia.org/wiki/Amidar})} The protocol matters for intervention because, for a lookup table, moving enemies will have no effect: enemies will simply ``teleport'' to the next location in the lookup table. 

The upper left plot in Fig.~\ref{fig:amidar-test-single} shows a baseline test for how an individual trained agent performs under each of four different enemy movement protocols: (1) a lookup table, on which the model was trained, (2) the ``Amidar movement'' protocol, (3) a random protocol, where at each junction the enemy chooses a random direction, and (4) an adversarial protocol, where enemies explore via random turns until the player is within line of sight, at which time they move toward the player's location. Note that, since enemies start far away from the player, the agent can (and does) easily make progress at the start of the game, regardless of enemy position. However, as the game progresses, the enemies close in and there are fewer opportunities for rewards.

The upper right plot in Fig.~\ref{fig:amidar-test-single} shows a test in which enemies ``gang up'' on the player: the enemies start position is modified to be close to the player. We were at first surprised to see how well the agent did; however, upon examination, we found we found that the agent was using up the jump button, which allows the player to bypass enemies, at beginning of the game. The lower half of Fig.~\ref{fig:amidar-test-single} depicts the results of running the baseline and test for no jumps: while the baseline performs similar, the player dies quickly for all non-lookup table enemy protocols.

\begin{figure}
    \centering
    \includegraphics[width=0.9\columnwidth]{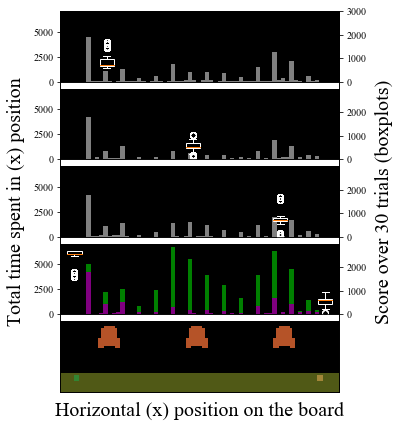}
    \caption{\label{fig:space-invaders-test}\textbf{Space Invaders case study.} The top three bands depict the tests of the solitary first, second, and third shields respectively. The bars depict the total number of steps spent in the corresponding horizontal location, while the boxplots depict the scores. The fourth band shows the behavior for all shields present (green; score boxplots on the left) and no shields present (purple; score boxplots on the right). The bottom band shows the Space Invaders terrain and default positions of the shields.  
    }
\end{figure}

\paragraph{Space Invaders: Shield Usage.} 
In Space Invaders, the player can seek refuge under three shields from the frontier of alien ships shooting down. We ran a test to see whether removing two of the three shields would cause an agent to use the remaining one more often. We also ran two baseline comparisons for a fixed amount of time: one where all sheilds are present (the default setting) and one where no shields were present.

Figure~\ref{fig:space-invaders-test} shows the results under test. Since score provides an incomplete picture of agent behavior, we also tracked the agent's location (a simple query in \toybox{}).  We observe that the player does not appear to change its preferred locations under any of the tests. 



\emph{A Note on Negative Results:} Space Invaders, as we have implemented it, has turned out to be a fairly uninteresting game. Randomly selecting from the trimmed action set that OpenAI Gym allows can lead to fairly good performance. Furthermore, our implementation, which included both random and adversarial enemy behavior, led the agent's behavior to be invariant to randomness in enemy behavior. 


\textbf{Findings.} We have shown a range of interventions and queries possible with \toybox{}, all of which would be impossible to conduct using ALE. The interventions we demonstrated were designed to demonstrate the power of \toybox{}'s design and implementation, rather than to satisfy any particular RL research agenda. We were able to rapidly iterate on all of our experiments due to \toybox{}'s fast performance and its simple API for editing state. In addition to highlighting \toybox{}'s capacity for evaluating a single agent, we have shown how \toybox{} may be used to evaluate models, by comparing the post-training performance ranking under test.

\section{Related Work}
\label{sec:related}
We are hardly the first to suggest new or different benchmarks for deep RL~\cite{kansky2017schema, zhang2018natural, wang2019paired}. Four major qualities differentiate \toybox{} from prior work: (1) it is based on a widely used and accepted community standard (ALE); (2) results on ALE can be replicated and compared in \toybox{}, providing continuity to individual research trajectories; (3) a wide array of features of \toybox{} environments are intervenable; furthermore, a particular configuration is easily exported and can be shared as part to further replication efforts; and (4) an individual game may be modified to produce a family of games, leading to a potentially infinite number of environments per-game; for example, the injection of real-world images into the background of Breakout described in~\cite{zhang2018natural} would be trivial to implement in \toybox{}.

Recall the available interventions in the traditional RL research environment shown in Figure \ref{fig:rl-system}. Most existing work manipulates the state input to the agent (i.e., the agent's \emph{perception} of state), the reward function, or the agent's actions, e.g.:
\begin{itemize}[leftmargin=*]
    \item\textit{State input}: Injecting real-world data into the background of Atari 2600 games simulates non-random noise~\cite{zhang2018natural}. Skipping frames periodically is a critically important hyperparameter for tuning algorithms to play Atari~\cite{mnih2013playing, braylan2015frame, frameskip}. 
    \item\textit{Reward function}: Hybrid reward structures decompose the reward function, making it easier for some agents to learn particularly difficult games in the Atari suite~\cite{van2017hybrid}.
    \item\textit{Agent actions}: Sticky actions, human starts, and random starts are all methods for intervening on the agent's actions outside the normal parameters of $\epsilon$-greedy exploration~\cite{sutton1998reinforcement, mnih2013playing, bellemare13arcade, nair2015gorila}.
\end{itemize} 
These efforts can help combat overfitting, learning spurious correlations, or generally failing to make progress on a task. \toybox{} is orthogonal to such efforts.

We have introduced relevant citations throughout the paper. Here we highlight critical work that was not otherwise mentioned.

\textbf{Evaluation and Replication.} Recent investigations into in how the community handles the evaluation and replication of agent performance has exposed some serious challenges that the community needs to address~\cite{henderson2017deep, balduzzi2018reevaluating, clary2018variability, jordan2018benchmarks}. Environments such as \toybox{} and evaluations of the style presented in Section~\ref{sec:case-studies} are one possible way to ameliorate issues surrounding replication, robustness, and variability. 

\textbf{Adversarial RL.} Much work on adversarial RL focuses on exploiting decision boundaries~\cite{mandlekar2017adversarially}, adding nonrandom noise to state input for the purpose of altering or misdirecting policies~\cite{huang2017adversarial}, or introducing additional agents to apply adversarial force during training to produce agents with more robust policies in physics simulations~\cite{pinto2017robust}. 

\textbf{Saliency maps.} Saliency maps were developed as an insight into model behavior~\cite{simonyan2013deep}, but more recently have been put forth as tool for explainability~\cite{ greydanus2017visualizing}. We show that in at least one case, saliency maps can be misleading, due in part to bias exemplified by the researcher's models of the agent and the environment as shown in Fig.~\ref{fig:rl-system}. Experiments enabled by \toybox{} provide much more specific information and can disambiguate competing hypotheses about agent behavior.

\section{Discussion}
\label{sec:discussion}
This paper is a proof-of-concept for experimentation about the behavior of deep RL agents. However, there are many possible applications beyond this type of post-training testing:

\textbf{Rejection sampling/dynamic analysis.} One of the biggest strengths offered in \toybox{} is the ability to answer arbitrary questions about the environment structures and code at any time. Agents may encounter local minima during training that are not representative of the target deployment distribution, due to factors such as random seeds~\cite{rlblogpost}. Training replicates with many random seeds is a costly solution to this problem. Instead, researchers could use \toybox{} to monitor environmental features to test whether an agent is spending too much time in an undesirable state.
Similar types of model monitoring could be used to identify ``detachment,'' a condition of the agent-environment interaction that induces catastrophic forgetting~\cite{kirkpatrick2017overcoming,goexplore}.

\textbf{Datasets from game families.} The ability to generate a family of games with similar but different mechanics provides a convenient dataset which can be used in a variety of ways. For example, with \toybox, a researcher can define a family of Breakout-style games with slightly different movement physics (e.g., ball velocity and acceleration, paddle-bounce mechanics) sampled from some real-valued parameter domain. This can be used to create a train/test split over environments~\cite{cobbe2018quantifying}, for supporting transfer learning experiments from one game physics to another~\cite{taylor2009transfer}, or for testing generalization across multiple environments~\cite{guestrin2003generalizing}.

\textbf{Adversarial testing.} With total control over environment dynamics, trained agents can be stress-tested by running the agent on progressively more difficult versions of the game. Tests of this form can serve to disambiguate agent behavior that can be explained in multiple ways---much as \toybox{}'s more advanced Amidar movement protocols revealed agents had not necessarily learned to avoid enemies so much as memorize their observed paths. Difficulty can be increased by increasing stochasticity in the environment, or increase the speed or accuracy of adversarial game elements. Similar methods could be used to create a suite of curriculum learning environments.

\section{Conclusions}
\label{sec:conclusion}
We have shown that \toybox{} unlocks novel and important capabilities for evaluating deep reinforcement learning agents. We introduce a new paradigm for thinking about evaluating agents, in the style of acceptance testing. We demonstrate \toybox{} capabilities with four case studies and outline a variety of other applications.


\bibliography{cites}
\bibliographystyle{icml2019}

\end{document}